\documentclass[conference, a4paper]{IEEEtran}
\IEEEoverridecommandlockouts
\usepackage{cite}
\usepackage{amsmath,amssymb,amsfonts}
\usepackage{graphicx}
\usepackage{textcomp}
\usepackage{xcolor}
\usepackage{dsfont}
\usepackage{tabulary,booktabs}

\usepackage{hyperref}
\usepackage{environ}
\usepackage{float} 
\usepackage{makecell}  
\usepackage{booktabs}
\usepackage{mathtools}
\usepackage{cancel}
\usepackage{verbatim}
\usepackage{dsfont}
\usepackage{subfigure}

\usepackage[ruled,linesnumbered]{algorithm2e}
\usepackage{float}
\usepackage{subcaption}

\makeatletter
\newcommand\newtag[2]{#1\def\@currentlabel{#1}\label{#2}}
\makeatother

\def\BibTeX{{\rm B\kern-.05em{\sc i\kern-.025em b}\kern-.08em
    T\kern-.1667em\lower.7ex\hbox{E}\kern-.125emX}}
    
\begin{document}

\title{FlatNAS: optimizing Flatness in Neural Architecture Search for Out-of-Distribution Robustness}

\author{\IEEEauthorblockN{Matteo Gambella, Fabrizio Pittorino, and Manuel Roveri}
\IEEEauthorblockA{Dipartimento di Elettronica, Informazione e Bioingegneria,\\
Politecnico di Milano, Milan, Italy\\
Email: \{matteo.gambella, fabrizio.pittorino, manuel.roveri\}@polimi.it} 
}

\maketitle


\begin{abstract}
Neural Architecture Search (NAS) paves the way for the automatic definition of Neural Network (NN) architectures, attracting increasing research attention and offering solutions in various scenarios. This study introduces a novel NAS solution, called Flat Neural Architecture Search (FlatNAS), which explores the interplay between a novel figure of merit based on robustness to weight perturbations and single NN optimization with Sharpness-Aware Minimization (SAM).
FlatNAS is the first work in the literature to systematically explore flat regions in the loss landscape of NNs in a NAS procedure, while jointly optimizing their performance on in-distribution data, their out-of-distribution (OOD) robustness, and constraining the number of parameters in their architecture.
Differently from current studies primarily concentrating on OOD algorithms, FlatNAS successfully evaluates the impact of NN architectures on OOD robustness, 
a crucial aspect in real-world applications of machine and deep learning. 
FlatNAS 
achieves a good trade-off between performance, OOD generalization, and the number of parameters, by using only in-distribution data in the NAS exploration. 
The OOD robustness of the NAS-designed models is evaluated by focusing on robustness to input data corruptions, using popular benchmark datasets in the literature.
\end{abstract}


\section{Introduction}%
\label{sec:introduction}

Deep learning models are widely spread nowadays~\cite{LeCun2015DeepL}. However, the
definition of deep neural network (NN) architectures is typically a complex and time-consuming task~\cite{automl}, often requiring a high expertise in the field~\cite{automl}. Neural Architecture Search (NAS)~\cite{nasframework}, a promising method of AutoML~\cite{automl}, aims at automating the design of a NN exploring different architectural configurations given a NN topology, a task to be accomplished, and a dataset used for training and validation to provide the optimal architecture.

Most of works on NAS focus on the design of NN architectures while taking into account target hardware. NAS solutions in this field fall into the area of Hardware-Aware NAS~\cite{HardwareNAS}. An emerging direction of these NAS solutions regards the use of constraints in the NAS exploration~\cite{gambella_cnas_2022}.
A constrained exploration is crucial in relevant scenarios such as TinyML~\cite{Roveri} and privacy-preserving deep learning solutions~\cite{gambella_cnas_2022}.


The traditional workflow of NAS, in both the constrained and non-constrained settings, deals with data that are sampled from a single dataset, resulting in NN architectures that are trained or fine-tuned on a single data distribution.
However, in real-world scenarios of modern machine and deep learning, mismatches of test and training data distributions are often observed, with deep learning models encountering significant performance drop in Out-Of-Distribution (OOD) scenarios~\cite{benchmarking_ood, hendrycks2021pixmix}.
This property undermines the trustworthiness of systems depending on such models and potentially threatens the safety of their users.
A crucial challenge in this direction is how much NN architectures are robust to changes on the distribution on which they are trained.
Out-of-distribution robustness considers
how to design the NN architectures resilient to various forms of
data shift at test time, with extensive efforts being devoted to improving OOD robustness \cite{pytorch-ood} through algorithmic enhancements. 
A natural algorithmic improvement for OOD robust NN architectures is to use NNs optimizers that look for flat regions in the loss landscape of NNs, such as SAM and its variants~\cite{sharpnessaware, asam, wsam}, as flatness-enhancing algorithms have been shown to optimize robustness to weight perturbation~\cite{pittorino2021}.
In fact, the geometrical structure of the loss landscape of NNs has been a key topic of study for several decades \cite{hochreiter, pittorino2021, annesi} and there is accumulating evidence of the connection between the flatness of minima found by optimization algorithms like stochastic gradient descent (SGD) and the generalization performance of the NN~\cite{baldassi2019shaping, jiang2019fantastic, lucibello_2022, pittorino22a}. Some works concentrate on the role of flatness-optimizing algorithms for OOD generalization~\cite{samood, swad, invlearning}, highlighting that this is a promising direction for algorithmic improvement.
As it is difficult to define principled guidelines for designing more robust NNs architectures, 
NAS is a natural candidate for automating their design.
Many works suggests that the existing NAS solutions, and more in general AutoML methods, search for models achieving optimal performance, while neglecting their robustness~\cite{security_automl,nasood}.
Furthermore,~\cite{empiricalstudyrobustness} explores the correlation between model complexity and robustness, highlighting that in a given architectural family, increasing the number of network parameters can lead to enhanced robustness. This suggests that networks constrained by the number of parameters are especially prone to degradation when faced with OOD data. 
These considerations highlight the need for a unified NAS framework that simultaneously optimizes all relevant aspects along these dimensions, i.e., increase the accuracy, increase the robustness, and reduce the number of parameters of the NN.
Currently, the NAS solutions addressing robustness are mainly designed to optimize the adversarial robustness of the NNs~\cite{nadar,racl,rnas,robnet,advrush,neighborhood-aware}, which is a somehow related but different task with respect to OOD robustness. In particular, some of these NAS algorithms account for the flatness of the input loss landscape of the network~\cite{advrush,neighborhood-aware}. 

The goal of our work is to develop Flat Neural Architecture Search (FlatNAS), a Neural Architecture Search algorithm, extending Constrained Neural Architecture Search (CNAS) \cite{gambella_cnas_2022}  by exploiting its constrained exploration while searching for flat regions in the loss landscape of NNs for enhanced generalization capabilities and OOD robustness. In particular, the explored architectures are constrained in their number of parameters (this case is of particular interest as smaller NNs are thought to be in general less robust~\cite{empiricalstudyrobustness}). In order to effectively account for flat regions, we introduce a novel metric, $R(x,\sigma)$, which assesses the capability of a given NN architecture $x$ to maintain accuracy under parameter perturbations of intensity $\sigma$, while optimizing single NN with SAM\cite{sharpnessaware, asam} instead of traditional SGD. We highlight that our robustness metric pertains to OOD robustness rather than adversarial robustness, and that our approach aims to design NNs with superior performance in OOD tasks by using exclusively in-distribution data in the NAS search process. 
The novel contributions of our research are twofold:
\begin{itemize}
\item the development of the first NAS algorithm specifically tailored to investigate flat regions within the optimization landscape of NNs with constrained number of parameters;
\item the introduction of a new metric designed to evaluate the robustness of NN architectures during NAS optimization.
\end{itemize}

The rest of this paper is organized as follows. Section~\ref{sct:related_literature} introduces the related works regarding NAS solutions for robustness. Section~\ref{sct:background} provides the background. 
Section~\ref{sct:tropicanas} introduces the FlatNAS framework developed in this study, while in Section~\ref{sct:experiments} the experimental results aiming at evaluating the effectiveness of FlatNAS are shown. Conclusions are finally drawn in Section~\ref{sct:conclusions}. To facilitate comparisons and reproducibility, the source code of FlatNAS is released to the scientific community as a public repository. \footnote{https://github.com/AI-Tech-Research-Lab/CNAS}

\section{Related literature}
\label{sct:related_literature}

This section explores NAS solutions enhancing the robustness of NN architectures. 
We initially review the existing body of work in NAS that addresses adversarial robustness. Then, we focus on studies that optimize the flatness of the loss landscape with respect to input data. Finally, we present the first example of NAS specifically tailored for OOD generalization. 

 In the field of NAS addressing adversarial robustness, RNAS~\cite{rnas} is a NAS solution derived from DARTS~\cite{liu_darts_2019} designed to optimize the balance between accuracy and adversarial robustness by implementing a regularization process that computes the similarity between outputs from natural and adversarial data. For computational efficiency, RNAS employs noise examples as proxies for adversarial data during the NAS search phase. 
 Differently, ROBNET~\cite{robnet}, utilizes an Once-For-All (OFA) supernet~\cite{ofa} in order to identify NNs that exhibit optimal adversarial robustness. It achieves this goal by sampling architectures from the supernet and subsequently fine-tuning these candidate architectures over a few epoch, to evaluate their accuracy when subjected to adversarial attacks.
 Similarly, NADAR~\cite{nadar}, based on DARTS, introduces a constrained optimization approach aimed at enhancing robustness within a specified computational budget measured in floating point operations (FLOPS). This method expands the backbone networks by integrating dilation networks, which are additional layers designed to augment robustness, considering their resulting complexity overhead. Finally, RACL~\cite{racl} presents a DARTS-based technique for optimizing robustness through a regularization process approximating the Lipschitz constant derived from the architecture parameters.

A different perspective is provided in ADVRUSH~\cite{advrush} and NA-DARTS~\cite{neighborhood-aware} whose goal is to address the problem of finding architectures with inherent robustness by targeting a smooth input loss landscape. More specifically, ADVRUSH, derived from DARTS, features an objective function that combines top1 accuracy on clean validation data and a smoothness-based regularization term. This regularization aims to optimize the curvature degree (flatness) of the input loss landscape evaluated using the largest eigenvalue of the Hessian matrix.  In contrast, NA-DARTS, another DARTS-based approach, differs from previous methods by assessing robustness through alterations in NN architecture, such as substituting a convolution with a skip connection. This solution aims to optimize the flatness of the NN architecture by employing an objective function focused on performance across the NN architecture of neighboring configurations.

Finally, NAS-OOD~\cite{nasood} is the first example of NAS method designed specifically for OOD generalization.  
Given a predefined set of source domains, the aim of the OOD task is to discover the optimal network architecture
that can generalize well to the unseen target domain. This DARTS-based approach optimizes the architecture on its performance on synthetically generated OOD data. A data generator is trained to create novel domain data by maximizing losses across various neural architectures, with NAS objective being to find parameters that effectively minimize these losses. The data generator and the neural architecture undergo simultaneous optimization. 
We highlight that the generator is designed to create images with different background patterns, textures, and colors. Differently, in our work we do not make any assumption about the type of perturbations considered and perform the NAS search only using in-distribution data. This study considers a general OOD generalization task and is not concentrated on OOD robustness to image corruptions and perturbations. Also, our method could be integrated with the proposed data generator.

\section{Background}
\label{sct:background}
\subsection{Neural Architecture Search}
\label{subsct:nas}

NAS solutions can be classified based on three distinct dimensions~\cite{nas_survey_components}.

The first dimension is the \textit{Search Space}, defining the architecture representation. The most straightforward is the \textit{entire-structured Search Space}, depicted layer-wise with each node representing a layer. Motivated by handcrafted architectures with repeated motifs~\cite{mobilenetv3}, the \textit{cell-based Search Space} defines each node as a cell (also called a block), representing a group of layers. Other variations include hierarchical cells or morphism-based identity transformations between layers.

The second dimension is the \textit{Search Strategy}, determining how to explore the search space. Common strategies include reinforcement learning, gradient optimization, and evolutionary algorithms
where genetic methods, especially NSGA-II~\cite{msunas}, are widely used. NSGA-II generates offspring using a specific crossover and mutation, selecting the next generation based on fitness through nondominated sorting and crowding distance comparison. It is a multi-objective algorithm, optimizing different figures of merit and effectively handling constraints~\cite{nsgaiiconstraints}, that can be managed by redefining the objective function. The simplest scheme is:
\begin{equation}
\label{eq:constraints}
\phi(x) = f(x) + pG(x)
\end{equation}
where $\phi(x)$ is the joint objective, \textit{f(x)} is the original objective without constraints, \textit{p} is the penalty term, and \textit{G(x)} is a function deciding whether to apply the penalty by accounting for the constraints. The penalty can be static (a constant value) or dynamic (adapted during the evolutionary process).

The third dimension is the \textit{Performance Estimation Strategy}, which estimates the performance without fully training every NN architecture in the search space (that would be computationally prohibitive). A popular strategy is weight sharing, with Once-For-All (OFA)~\cite{ofa} being a state-of-the-art example. OFA trains a comprehensive supernet of many network configurations once and evaluates candidate networks by fine-tuning from the supernet weights during NAS. The OFA supernet is trained, before the NAS search, using Progressive Shrinking, starting with the largest NN and progressively fine-tuning to support smaller sub-networks sharing weights. Another approach involves surrogate models such as Gaussian Process
and Radial Basis Function
For instance, MSuNAS~\cite{msunas} proposes \textit{adaptive-switching}, selecting the best accuracy predictor among four surrogate models based on a correlation metric named Kendall’s Tau
in each NAS iteration. 

\subsection{Sharpness-Aware Minimisation (SAM)}
\label{subsct:sam}

Sharpness-Aware Minimization (SAM) has emerged as a significant advancement in the field of machine learning, particularly for deep neural networks, thanks to its ability to reduce generalization error by minimizing a sharpness measure that reflects the geometry of the loss landscape~\cite{sharpnessaware}. Traditionally, training of DNNs has focused on minimizing empirical loss, often leading to overfitting and convergence to sharp minima~\cite{pittorino22a}. SAM addresses this issue by seeking flat minima through a min-max optimization problem, involving two forward-backward computations for each update. This approach achieved remarkable results in training various deep neural networks~\cite{zhuang2022surrogate, wsam}.

\section{The proposed FlatNAS}
\label{sct:tropicanas}

This section, introducing and detailing FlatNAS, is organized as follows:
Section~\ref{subsct:problem_formulation} provides the problem formulation,
Section~\ref{subsct:overallview} provides an overall description of FlatNAS,
Section~\ref{subsct:top1_robust} presents the figure of merit accounting for classification accuracy and robustness on OOD generalization, and finally
Section~\ref{subsct:constr_params} introduces the figure of merit accounting for the number of parameters and the related constraint.

\subsection{Problem formulation}
\label{subsct:problem_formulation}
FlatNAS addresses the challenge of selecting a NN architecture by simultaneously optimizing classification accuracy and robustness to weight perturbations, while satisfying a constraint on the maximum number of parameters. FlatNAS can be defined as the following optimization problem:
\begin{align}
\label{eq:constrained_problem}
    \text{minimize }  &\mathcal{G} \left ( 
    F_{A}(x), R(x, \sigma), F_P(x) \right ) 
    \nonumber \\
    \text{s. t. } & F_P(x) < \overline{F}_P, \\
    &x \in \Omega_{x} \nonumber
\end{align}
where $\mathcal{G}$ is a multi-objective optimization function, $x$ and $\Omega_{x}$ represent a candidate NN architecture and the search space of the NN exploration, respectively; the metrics $F_{A}(x)$ and $R(x, \sigma)$ calculate the top-1 accuracy and the robustness of the architecture $x$, respectively. $F_P(x)$ represents the number of parameters in architecture $x$, and $\overline{F}_P$ denotes the upper limit on the allowable number of parameters. The optimization problem outlined in Eq. \eqref{eq:constrained_problem} is addressed by the FlatNAS framework, which is detailed in what follows.

\subsection{The overall view of FlatNAS}
\label{subsct:overallview}
\begin{figure*}[h!]%
	\centering
	\includegraphics[width=0.8\linewidth]{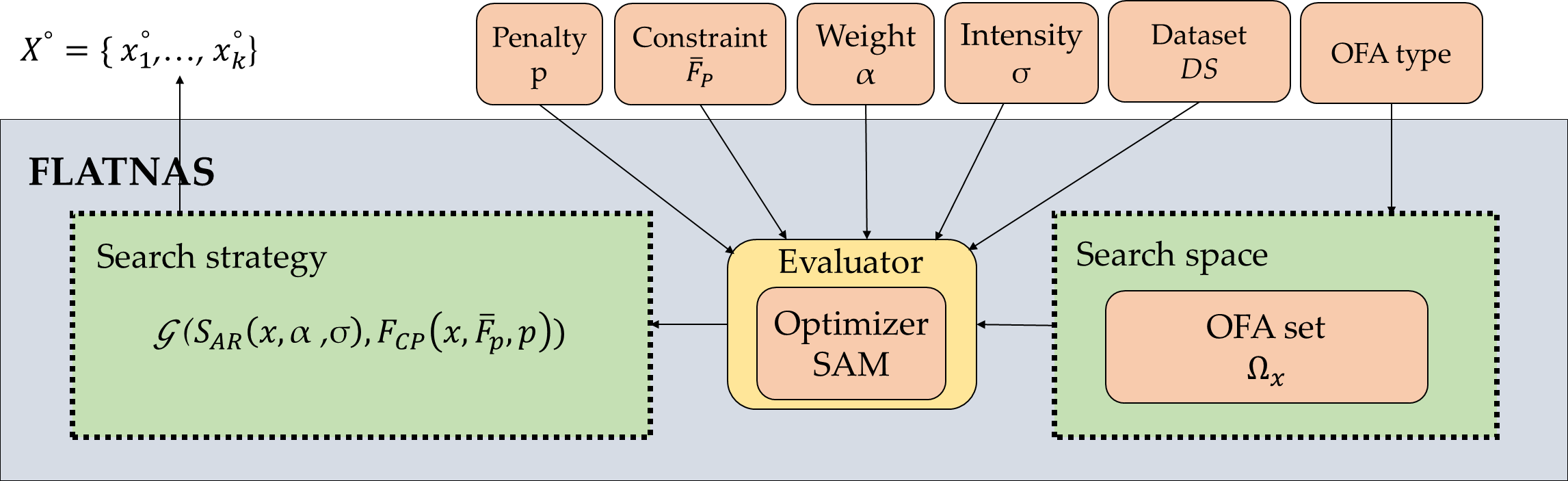}
	\caption{The proposed FlatNAS framework, which is composed of a Search Space, Evaluator, and a Search Strategy module.} 
	\label{fig:tropicanas_scheme}
\end{figure*}

The FlatNAS framework, illustrated in Fig.~\ref{fig:tropicanas_scheme}, operates as follows. It receives a dataset $\mathcal{DS}$, which includes a training set for training candidate networks and a validation set for their validation. The framework utilizes a type of OFA supernet to select a set of candidate networks $\Omega_x$, defining the \textit{Search Space} of NAS. Additionally, FlatNAS incorporates the constraint $\overline{F}_P$ and the associated penalty in the objective function optimized by the \textit{Search Strategy} to limit the number of parameters. Integral components of the framework are the intensity of the perturbation $\sigma$, used by the \textit{Evaluator} to assess robustness, and the weight $\alpha$, used to combine classification performance and robustness.

The \textit{Evaluator} module receives a NN architecture from the set $\Omega_{x}$ and trains it by using the $\mathcal{DS}$ dataset and the SAM optimizer. Subsequently, it assesses the NN robustness. The resulting output is an \textit{archive}, i.e., a collection of tuples $\langle x, F_A(x), F_P(x), R(x,\sigma) \rangle$, representing the NN configurations and their corresponding figures of merit, i.e., $F_A(x)$ for accuracy, $F_P(x)$ for the number of parameters, and $R(x,\sigma)$ for robustness. Initially, a representative subset of OFA $\Omega_{x}$ is sampled before starting the search. The number $N_{start}$ of selected NN architectures is determined by the user and it is typically set to $100$. Therefore, the initial size of the archive is equal to $N_{start}$.

In FlatNAS, the optimization problem outlined in Eq. \eqref{eq:constrained_problem} is reformulated to implicitly incorporate constraints:
\begin{align}
\label{eq:tropicanas_problem}
    \text{minimize }  &\mathcal{G} \left ( 
    S_{AR}(x,\alpha,\sigma), F_{CP}(x, \overline{F}_P, p) \right )   \\
    \text{s. t. }  & x \in \Omega_x
    \nonumber
\end{align}
where $\mathcal{S_{AR}}(x, \alpha, \sigma)$ is the predicted value of $F_{AR}(x, \alpha, \sigma)$ as estimated by a surrogate model, and $F_{CP}(x,\overline{F}_P,p)$ accounts jointly for the number of parameters of the architecture and the related constraint. In particular, $F_{AR}(x, \alpha, \sigma)$ is a novel figure of merit, which will be introduced in Section~\ref{subsct:top1_robust}, accounting for both the classification accuracy and the robustness of model $x$. It utilizes a weight $\alpha$ to balance the importance of these two factors with the intensity of the perturbation governed by the parameter $\sigma$.

The core of FlatNAS is the \textit{Search Strategy} module. It adopts the NSGA-II genetic algorithm~\cite{msunas} to solve the bi-objective problem introduced in Eq. \eqref{eq:tropicanas_problem}, by optimizing the objectives $S_{AR}(x,\alpha,\sigma)$ and $F_{CP}(x,\overline{F}_P,p)$. The search process is iterative: at each iteration, the surrogate model, computing $S_{AR}(x,\alpha,\sigma)$, is chosen employing a mechanism called \textit{adaptive-switching}, the same method used by MSuNAS~\cite{msunas}, which selects the best surrogate model according to a correlation metric (i.e., Kendall's Tau).
The surrogate models are trained by using the \textit{archive} as the dataset. We highlight that $F_{CP}(x,\overline{F}_P,p)$ is computed analytically as shown in Section~\ref{subsct:constr_params}, hence a surrogate model is not needed. Then, a ranking of the candidate NNs, based on $S_{AR}(x,\alpha,\sigma)$ and $F_{CP}(x,\overline{F}_P,p)$, is computed and a new set of candidates is obtained by the genetic algorithm and forwarded to the \textit{Evaluator}. The \textit{Evaluator} updates the \textit{archive}, which becomes available for evaluation in the next iteration.
At the end of the search, FlatNAS returns the set of the $k$ NN architectures $X^\circ = \{ x_1^\circ, \ldots, x_k^\circ \}$ characterized by the best trade-off among the objectives, where $k$ is a user-specified value. 

\subsection{The figure of merit $F_{AR}(x, \alpha, \sigma)$}
\label{subsct:top1_robust}

In this section, we define the novel figure of merit of FlatNAS accounting jointly for the accuracy and the robustness.

To achieve this goal, we use a rescaling-invariant notion of flatness used in \cite{jiang2019fantastic, pittorino2021}. Given a weight configuration $w(x)\in \mathbb{R}^N$ (i.e., the weight relative to architecture $x$), we define the \emph{robustness} $R(x, \sigma)$ as the average training error difference with respect to $E_\text{train}(w(x))$ when perturbing the weight configuration $w(x)$ by a (multiplicative) noise proportional to a parameter $\sigma$, i.e.,
\begin{equation}
    R(x, \sigma) =\mathbb{E}_z\, E_\text{train}(w(x)+\sigma z \odot w(x))
     - E_{\text{train}}(w(x)),
\label{eq:localE}
\end{equation}
where $\odot$ denotes the element-wise product and the expectation is over normally distributed $z\sim \mathcal{N}(0, I_N)$. In our experiments we set $\sigma=0.05$ (in principle varying the parameter $\sigma$ can be informative~\cite{pittorino2021}, however a single value already shows a good correlation with generalization~\cite{jiang2019fantastic}). 
In practice, we compute $R(x, \sigma)$ as the average value of $E_\text{train}(w(x)+\sigma z \odot w(x)) - E_\text{train}$ over a user-defined number of samples $M$ from~$z$ (in our case $M$ = 20). 

Finally, we define $F_{AR}(x,\alpha, \sigma)$ as the weighted sum of two different terms as follows:
\begin{equation}
    \label{eq:top1_robust}
    \begin{aligned}
        F_{AR}(x,\alpha, \sigma) &= \alpha (1-F_A(x)) + \gamma (1-\alpha) R(x, \sigma), \\
        \gamma &= \left. \left(\sum_{i=1}^{N_k}{F_A(x_i)}\right)\middle / \right. \left(\sum_{i=1}^{N_k}{R(x_i, \sigma)}\right)
    \end{aligned}
\end{equation}
where $F_A(x)$ is the top-1 classification accuracy on the validation set, while $\alpha$ is a user-defined parameter that controls the relative significance of optimizing the two components, and $\gamma$ is the ratio between the average value of the top1 accuracy and the average value of the robustness of the architectures in the \textit{archive} at iteration k of the NAS with $N_k$ the cardinality of the \textit{archive} at iteration k. This latter term balances the order of magnitude between the top1 accuracy and the robustness in order to let $\alpha$ work effectively.

\subsection{The figure of merit $F_{CP}(x, \overline{F}_P, p)$}
\label{subsct:constr_params}
In this section, we introduce the definition of the figure of merit $F_{CP}(x, \overline{F}_P, p)$ accounting for the number of parameters and the constraints related.

A viable approach to incorporate constraints into a figure of merit is the one used in~\cite{gambella_cnas_2022}. Given a constraint on the maximum number of parameters $\overline{F}_P$ and a penalty factor \textit{p}, a possible figure of merit $F_{CP}$ accounting for the constraint is:
\begin{equation}
    \label{eq:c_params}
      F_{CP}(x, \overline{F}_P, p) = F_P(x) + max(0,F_P(x)-\overline{F}_P)p
\end{equation}

The figure of merit $F_{CP}(x, \overline{F}_P, p)$ imposes a penalty when the number of parameters in a model exceeds the maximum allowable limit. Conversely, no penalty is applied (as indicated by the max operator returning zero) when the constraint is satisfied. The extent of the penalty is proportional to the degree of constraint violation and the value of the constant $p$.

\subsection{The use of Sharpness-Aware Minimization (SAM)}
\label{subsct:SAM}
In order to enhance flatness optimization at the single NN level, instead of SGD we use SAM~\cite{sharpnessaware} for optimizing NN architectures $x$ during the NAS exploration. 
We will see that the use of SAM has a non-trivial effect on the NAS architectures space exploration, resulting not only in an optimization of the figure of merit $F_{AR}(x,\alpha, \sigma)$ and consequently on more robust models on OOD data, but also in qualitatively different architectures. These aspects are detailed in the experimental results that are introduced in the next Section.

\section{Experimental results} 
\label{sct:experiments}
This section describes the experimental results 
aiming at assessing the effectiveness of FlatNAS. 

\subsection{Out-Of-Distribution (OOD) datasets}
\label{subsct:ood}
Numerous studies have shown that datasets with corrupted versions of natural data, such as images affected by noise or more complex distortions, can be utilized to evaluate model robustness~\cite{pytorch-ood, benchmarking_ood}.
In particular,~\cite{benchmarking_ood} introduces the first comprehensive benchmarks aimed specifically at robustness, concentrating on various types of corruption and perturbation.
The CIFAR-10-C and CIFAR-100-C benchmarks \cite{benchmarking_ood}, where the C at the end stands for the corrupted version of the original datasets, consist of a wide range of image corruptions that have been demonstrated to reflect robustness against certain real-world data variations.
These benchmarks are designed to challenge models trained on the CIFAR-10 and CIFAR-100 datasets by providing a more demanding, held-out test set. 
Our experiments employ the original, clean datasets CIFAR-10 and CIFAR-100,
along with their corrupted versions, CIFAR-10-C and CIFAR-100-C, which include 15 different types of distortions. Since the corrupted datasets are available only with a resolution of 32 while we consider NNs with different resolutions, we reproduced the corrupted data for each resolution evaluated. We highlight that simply resizing the images of the original corrupted datasets would not be equivalent.
More information about the specific distortions and an illustration of the transformations, can be found in~\cite{benchmarking_ood} and Fig.~\ref{fig:OOD}, respectively.

\subsection{Experimental setup} 

The setup of the experiments is the following:
\begin{itemize}
 \item the first objective is $F_{AR}(x,\alpha,\sigma)$ and the second objective is $F_{CP}(x,\overline{F}_P,p)$;
 \item the initial number of samples populating the archive $N_{start}$ is
set to 100 and the total number of NAS iterations is set to $30$ (at each iteration a set of $8$ new candidate NNs is added to the archive);
\item the OFA supernet is based on MobileNetV3 and the hyperparameters of $\Omega_x$ refer to resolution (range from 128 to 224 with step size 4),  depth (2, 3, or 4), kernel size (3, 5, or 7) and expansion rate (3, 4, or 6);
\item we use learning rate $0.1$, momentum $0.9$, batch size $128$ and weight decay $0.0005$ for the SGD optimizer;
\item we use an adaptive SAM optimizer~\cite{asam} with the same hyperparameters used for SGD, and parameter $\rho=2$ as empirically suggested in~\cite{asam}.
\end{itemize}

The set of optimal architectures $X^{\circ}$ from each NAS search contains the architecture with the best trade-off between $F_{AR}(x,\alpha,\sigma)$ and $F_{CP}(x,\overline{F}_P,p)$. We refer to these networks as \textit{FlatNAS}$\alpha$$X$$\sigma$$Y$, where $X$ is the value of $\alpha$, the weight of terms in $F_{AR}$, and $Y$ is the value of $\sigma$, the perturbation parameter used in computing $R(x,\sigma)$.

For comparative purposes, we compare FlatNAS with the basic version of CNAS, which uses SGD and the figure of merit $F_A(x)$, rather than SAM and the novel metric $F_{AR}(x)$. 
Taking CNAS as a baseline aims to assess the FlatNAS framework effectiveness in identifying robust and high-performing NNs with respect to standard NAS solutions that do not specifically enhance robustness.
\begin{figure}[ht]
	\centering
	\includegraphics[width=0.475\textwidth]{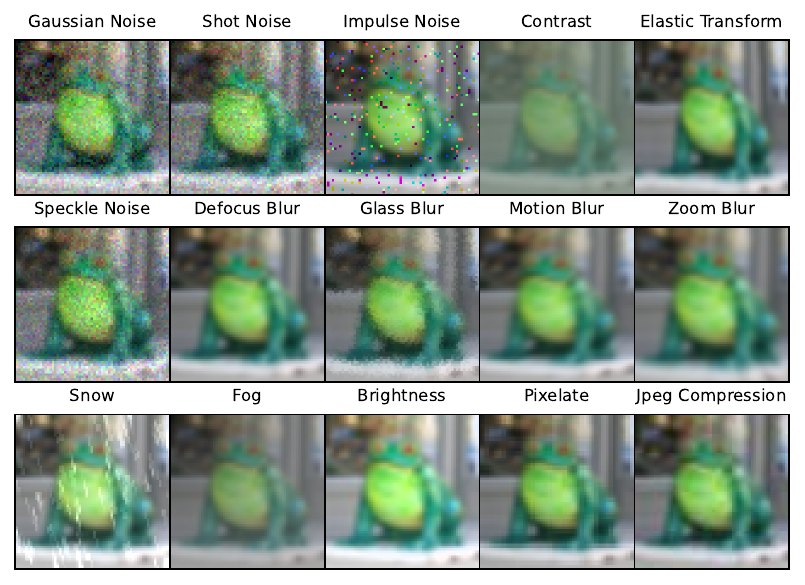}
	 \caption{\label{fig:corruptions} Image corruptions from the CIFAR-10-C dataset. The figure shows the 15 noise types with level of severity set to 3 with image resolution 64.
  }
\end{figure}

\subsection{Analysis of the results}
A summary of experimental results on CIFAR-10 and CIFAR-100 (and their respective corrupted versions CIFAR-10-C and CIFAR-100-C) is reported in Tables~\ref{tab:CIFAR-10}~and~\ref{tab:CIFAR-100}. In each table we show a comparison between FlatNAS at different $\alpha$-values and CNAS (trained with SGD), with respect to top1 accuracy $F_A(x)$, the robustness $R(x,\sigma)$, the figure of merit $F_{AR}$, the number of parameters (\textit{Params}), the number of multiply-accumulate operations (\textit{MACs}) and the mean Corruption Error (\textit{mCE}).
To ensure a fair benchmarking of CNAS against other FlatNAS networks on OOD datasets, we show additional results in which at the end of the CNAS exploration the resulting NN model is optimized with SAM. This step isolates the positive impact of the SAM optimizer, hence allowing us to more accurately evaluate the capability of FlatNAS in identifying the most OOD-robust architecture.

The \textit{mCE} metric that we report in Tables~\ref{tab:CIFAR-10}~and~\ref{tab:CIFAR-100} refers to the average value of the Corruption Error mediated over all corruption types and their intensities. As suggested in ~\cite{benchmarking_ood}, it is calculated as
$ \mathrm{CE}_{c}=\sum_{s=1}^{5} E_{s, c}$,
where $s$ is the severity level and $c$ the corruption type. In this way model corruption robustness is summarized by averaging the 15 Corruption Error values, i.e., $\mathrm{CE}=$ $\{\mathrm{CE}_{\text {Gaussian Noise }}, \mathrm{CE}_{\text {Shot Noise }}, \ldots, \mathrm{CE}_{\mathrm{JPEG}}\}$, resulting in the \textit{mean CE} (\textit{mCE}). We do not subtract from the \textit{mCE} the original classification error, which would result in the \textit{relative mCE}, as our goal is the \textit{mCE} over architectures.

\begin{figure}[ht]
	\includegraphics[width=0.47\textwidth]{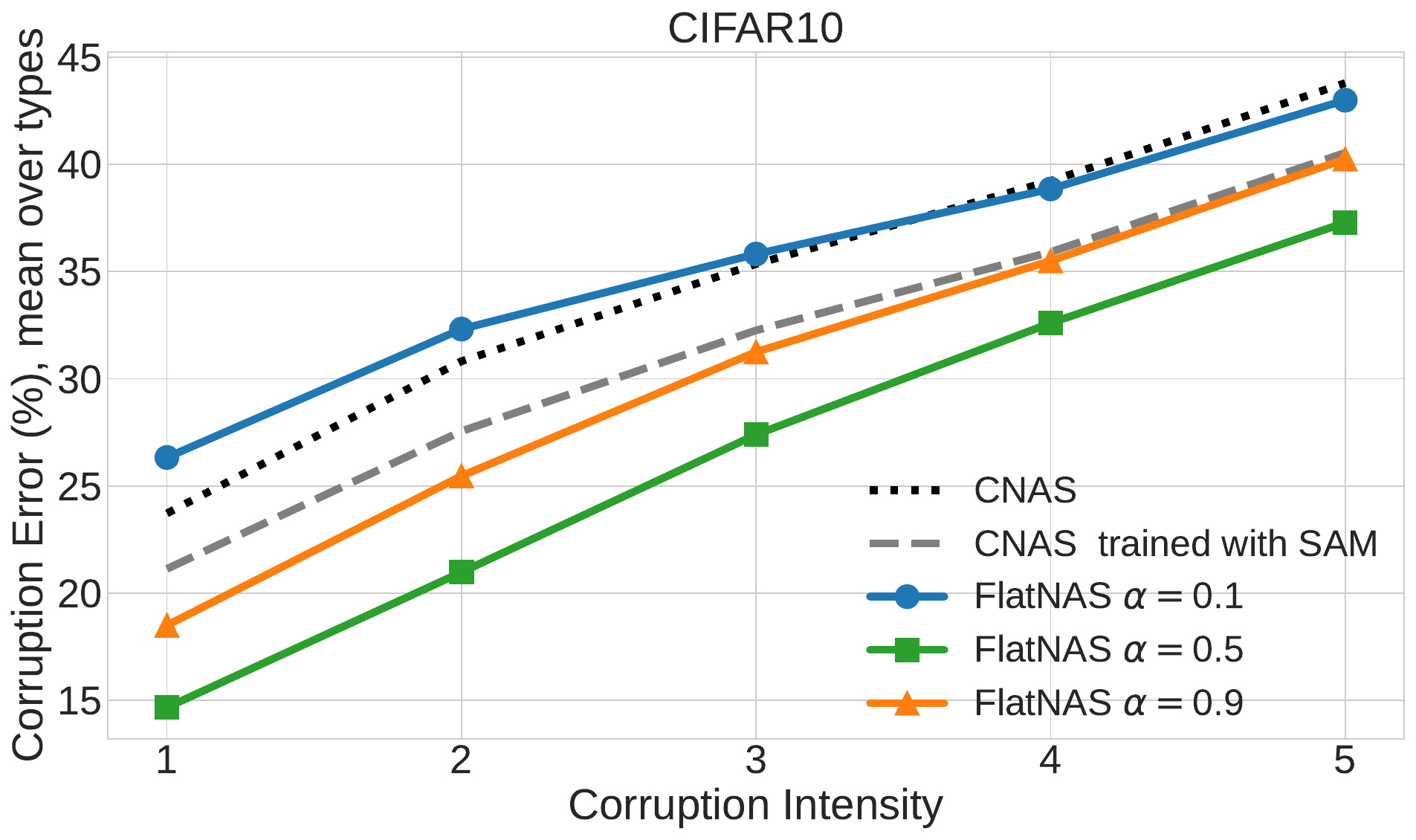}
	\includegraphics[width=0.47\textwidth]{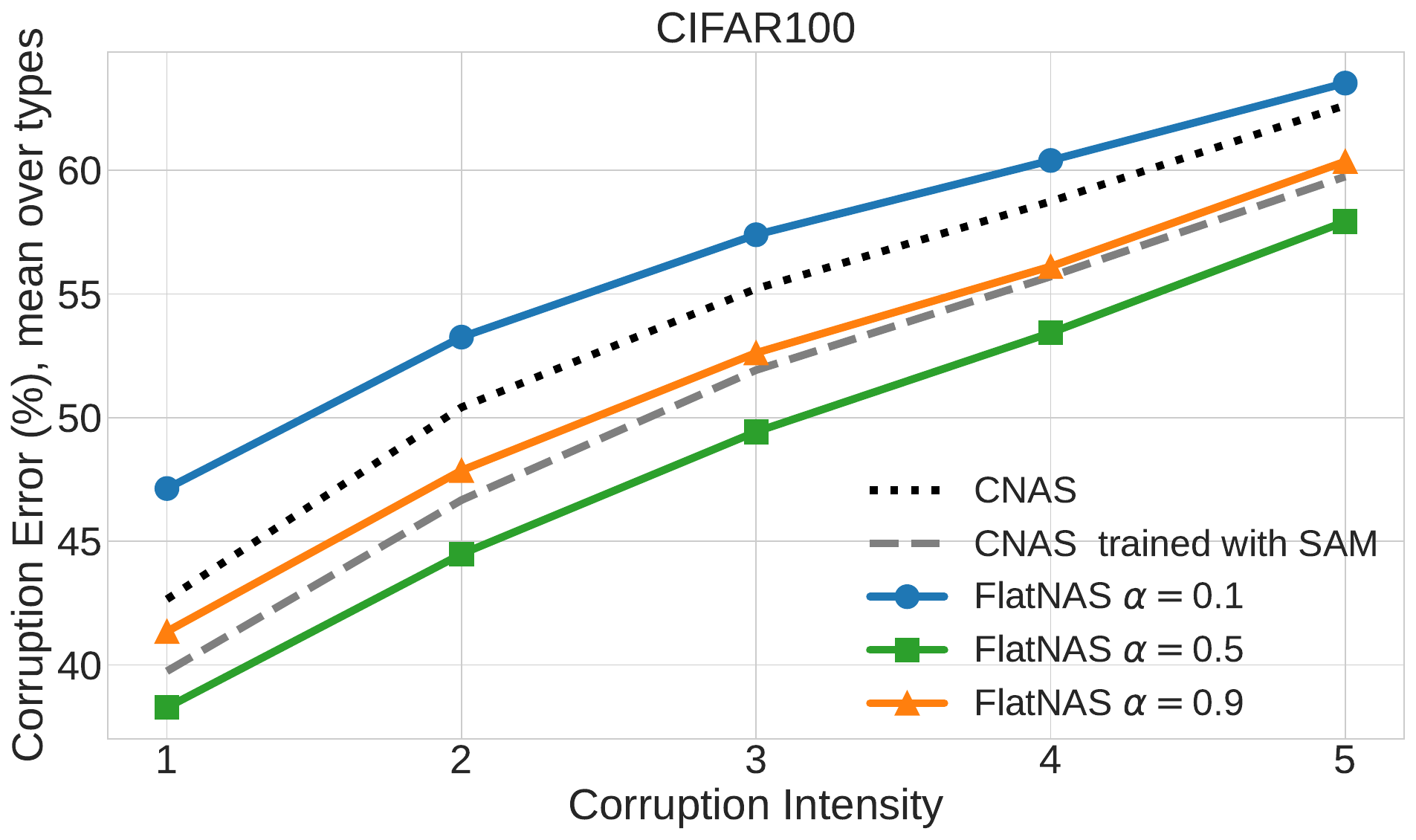}
	 \caption{\label{fig:OOD} Comparison of mean NN corruption errors on corrupted datasets in function of the perturbation intensity, for FlatNAS at different $\alpha$ values and CNAS. (Upper panel: NNs trained on CIFAR-10 and evaluated on CIFAR-10-C; lower panel: same as upper panel but for CIFAR-100.)
	 }
\end{figure}

\begin{figure*}[h!]
	\includegraphics[width=0.98\textwidth]{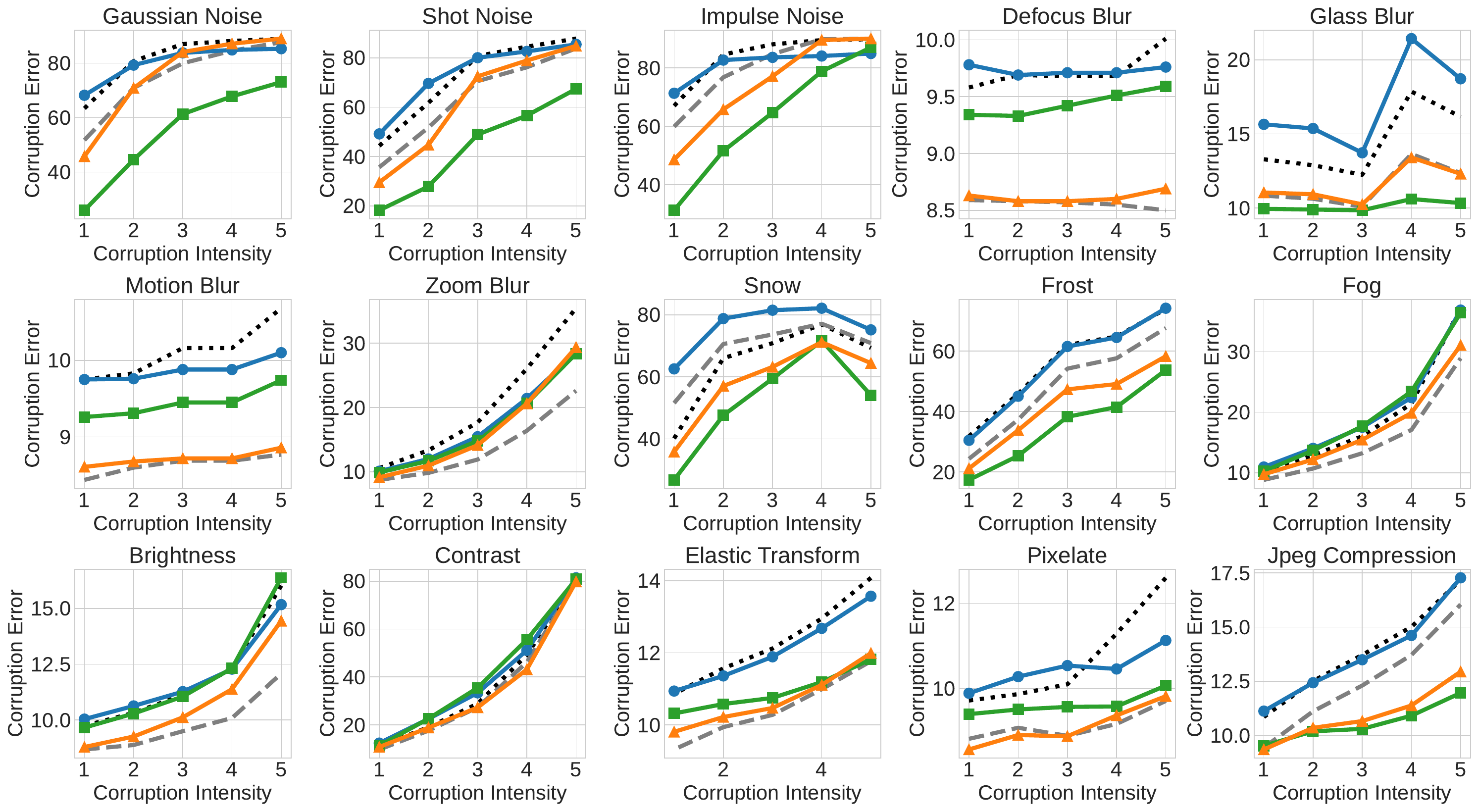}
	 \caption{\label{fig:OOD10} Final  generalization profiles on CIFAR-10-C of NNs trained on  CIFAR-10, in function of the perturbation intensity and for each corruption type. Dotted lines correspond to the best NN model found by CNAS and trained with SGD, dashed lines correspond to the latter NN model trained with SAM, and full lines correspond to the best model found by FlatNAS (the dot symbol corresponds to $\alpha=0.1$, the square to $\alpha=0.5$ and the triangle to $\alpha=0.9$).
	 }
\end{figure*}

\begin{figure*}[h!]
	\includegraphics[width=0.98\textwidth]{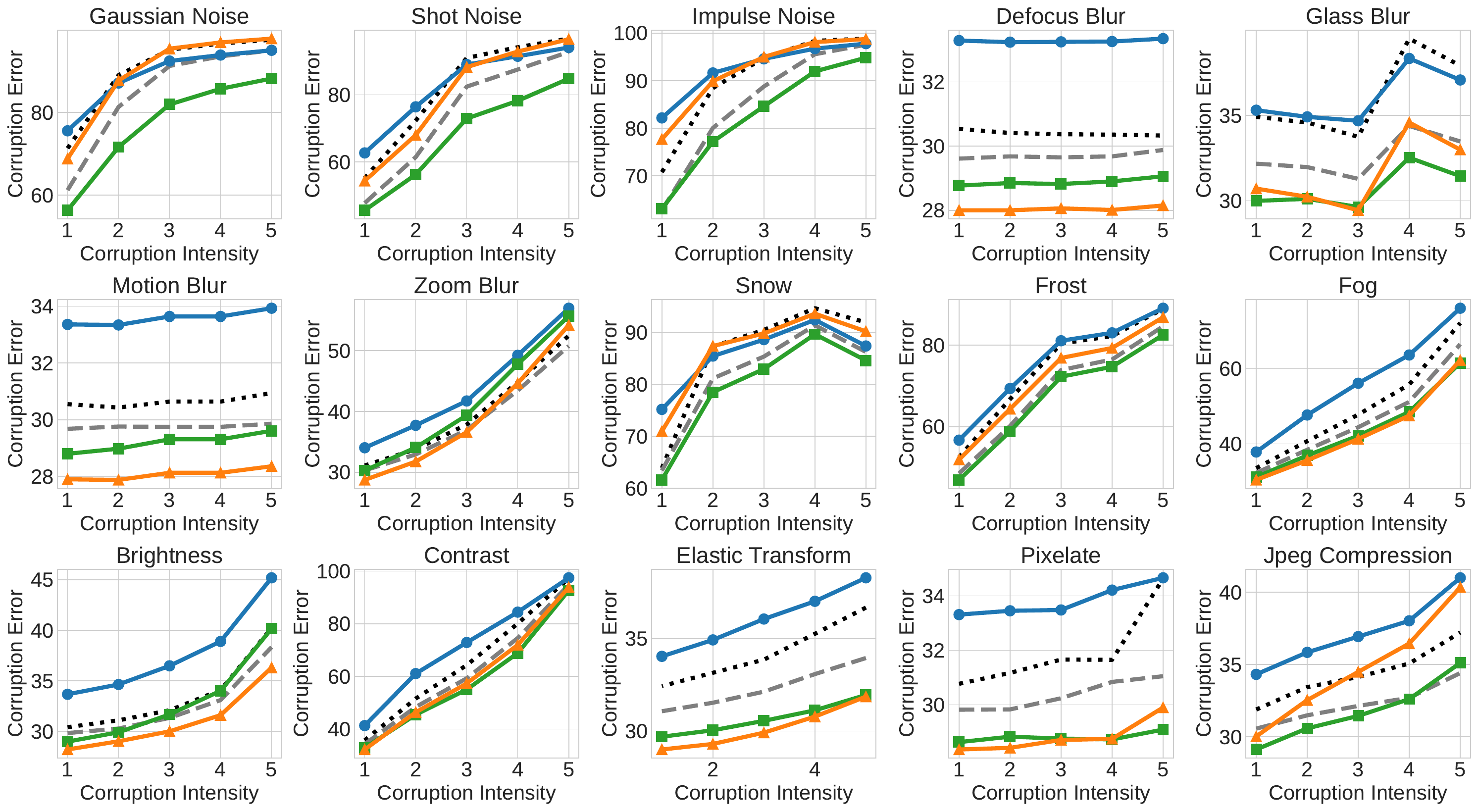}
	 \caption{\label{fig:OOD100} 
        Same as Fig.~\ref{fig:OOD10} but for CIFAR-100-C.
	 }
\end{figure*}

To show all the information that may be hidden in the single \textit{mCE} value, in Fig.~\ref{fig:OOD} we show the \textit{CE} mediated over all corruptions as a function of the intensity value, providing a good summary of the main results showcasing enhanced robustness for NN models found by FlatNAS with respect to CNAS. In Figs.~\ref{fig:OOD10}~and~\ref{fig:OOD100} we report a detailed comparison between the performance of the NNs models found by FlatNAS and CNAS for each distortion type present in CIFAR-10-C and CIFAR-100-C, as a function of the intensity value. 
\begin{table}[ht]
\centering
\caption{Results of FlatNAS and CNAS on CIFAR10
and CIFAR10-C.}
\label{tab:CIFAR-10}
\scalebox{0.9}{
\begin{tabular}{@{}c|c|c|c|c|c@{}}
\toprule
Model  & $1-F_A(x)$  & $R(x,\sigma)$ & mCE & Params & MACs  \\ \midrule
FlatNAS$\alpha0.1\sigma0.05$ & 9.81 & 7.85 & 35.26 & 5.63 & 261.90 \\ \midrule
FlatNAS$\alpha0.5\sigma0.05$ & 9.34 & 2.11 & 26.58 & 4.34 & 173.63 \\ \midrule
FlatNAS$\alpha0.9\sigma0.05$ & 8.62 & 6.18 & 30.19 & 4.40 & 267.94 \\  \midrule
CNAS (SAM) & 8.60 & 2.18  & 31.48 & 4.50 & 252.20 \\  \midrule
CNAS & 9.61 & 17.51  & 34.58 & 4.50 & 252.20 
\end{tabular}
}
\end{table}
\begin{table}[ht]
\centering
\caption{Results of FlatNAS and CNAS on CIFAR100 and CIFAR100-C.}
\label{tab:CIFAR-100}
\scalebox{0.9}{
\begin{tabular}{@{}c|c|c|c|c|c@{}}
\toprule
Model  & $1-F_A(x)$  & $R(x,\sigma)$ & mCE & Params & MACS   \\ \midrule
FlatNAS$\alpha0.1\sigma0.05$ & 31.62 & 8.42 & 56.35 & 4.72 & 254.66 \\ \midrule
FlatNAS$\alpha0.5\sigma0.05$ & 27.74 & 6.46 & 48.71 & 5.15 & 338.45 \\ \midrule
FlatNAS$\alpha0.9\sigma0.05$ & 27.24 & 6.8 & 51.66 & 6.85 & 376.22 \\  \midrule
CNAS (SAM) & 29.11 & 7.72  & 50.76 & 5.19 & 295.61 \\  \midrule
CNAS & 29.59 & 26.59 & 53.93 & 5.19 & 295.61
\end{tabular}
}
\end{table}

In Tables~\ref{tab:CIFAR-10}~and~\ref{tab:CIFAR-100}, and in Fig.\ref{fig:OOD}, we can see that FlatNAS is able to reduce significantly the \textit{mCE} values without incurring in a significant loss in the accuracy $F_A(x)$, while maintaining a comparable number of parameters, which is our principal result. Notice that this holds for every value of the corruption intensity. 
Meanwhile, NN models found by FlatNAS with $\alpha=0.9$ yield comparable results to models obtained with CNAS and then trained with SAM, indicating that in this case FlatNAS is not effective in finding the architecture with the highest robustness. With $\alpha=0.1$, $R(x,\sigma)$ is the main term optimized in $F_{AR}(x,\alpha, \sigma)$, resulting in NNs with worse $F_{A}(x)$ score, resulting in a degradation of the mCE. We see that in Figs.~\ref{fig:OOD10} and~\ref{fig:OOD100}, where we isolate the error on each corruption type, the general trend observed in Fig.\ref{fig:OOD} is respected. In particular, on the most difficult distortion (i.e., yielding high values of corruption errors), FlatNas achieves excellent improvements in robustness with respect to CNAS. For instance from Fig.~\ref{fig:OOD10} we see that on CIFAR-10-C, at the lowest corruption intensity value, on Gaussian Noise FlatNAS scores $26\%$ of corruption error and CNAS $63\%$, while on Impulse Noise they achieve $31\%$ and $67\%$ respectively.
In Fig.~\ref{fig:OOD} and Tables~\ref{tab:CIFAR-10}~and~\ref{tab:CIFAR-100} we also show results varying the $\alpha$ value in the figure of merit $F_{AR}(x,\alpha, \sigma)$ defined in Eq.~\eqref{eq:top1_robust}. We observe that $\alpha=0.5$ gives better mCE (and therefore OOD robustness). This indicates that both terms in $F_{AR}(x,\alpha, \sigma)$ are equally important for optimizing OOD robustness.

It is worth noting that NN models identified by FlatNAS are in general different with respect the ones found by CNAS. 
In particular, FlatNAS consistently employs a diverse range of kernel sizes (3, 5, 7) across both datasets, with a noticeable inclusion of larger kernels (7). CNAS also utilizes a range of kernel sizes but does not show as clear a pattern in the preference for larger kernels across datasets. 

\section{Conclusions}
\label{sct:conclusions}
In this work, we have addressed the timely issue of NAS for OOD robustness.
We have focused on parameter-constrained NNs, which are expected to particularly suffer the issue of robustness. Our results indicate that sharpness-aware methods are successful in enhancing OOD robustness in NAS, modifying in a non-trivial manner the NAS exploration and resulting in qualitatively different and more robust architectures. Future works will examine different OOD tasks, OFAs, application scenarios, and the role of architectural perturbations.

\bibliographystyle{IEEEtran}
\bibliography{IEEEabrv,main.bib}

\end{document}